# Approches d'analyse distributionnelle pour améliorer la désambiguïsation sémantique

Mokhtar Boumedyen Billami[1], Núria Gala[1]

[1]Aix-Marseille Université, LIF UMR 7279, Marseille – France

## Abstract

Word sense disambiguation (WSD) improves many Natural Language Processing (NLP) applications such as Information Retrieval, Machine Translation or Lexical Simplification. WSD is the ability of determining a word sense among different ones within a polysemic lexical unit taking into account the context. The most straightforward approach uses a semantic proximity measure between the word sense candidates of the target word and those of its context. Such a method very easily entails a combinatorial explosion. In this paper, we propose two methods based on distributional analysis which enable to reduce the exponential complexity without losing the coherence. We present a comparison between the selection of distributional neighbors and the linearly nearest neighbors. The figures obtained show that selecting distributional neighbors leads to better results.

## Résumé

La désambiguïsation sémantique permet d'améliorer de nombreuses applications en traitement automatique des langues (TAL) comme la recherche d'information, la traduction automatique ou la simplification lexicale de textes. Elle consiste à choisir le sens des unités lexicales polysémiques dans un texte et s'effectue en tenant compte du contexte. L'approche la plus directe consiste à estimer la proximité sémantique entre chaque sens candidat et les sens des mots du contexte. Cette méthode engendre rapidement une explosion combinatoire. Dans cet article, nous proposons deux approches à base d'analyse distributionnelle permettant de réduire la complexité exponentielle et de ne pas perdre de la cohérence au niveau de la désambiguïsation, cela en sélectionnant les voisins distributionnels les plus proches. Nous présentons une comparaison entre la sélection des voisins distributionnels et les voisins les plus proches linéairement. Les résultats montrent que la sélection des voisins distributionnels est bien meilleure.

**Mots clés:** désambiguïsation sémantique non supervisée, analyse distributionnelle, analyse syntaxique en dépendances, représentation vectorielle continue.

## 1. Introduction

La désambiguïsation des sens de mots est essentielle pour accomplir la plupart des tâches de traitement des langues (Navigli, 2009), par exemple, la recherche d'information, la traduction automatique, l'extraction d'information, l'analyse du contenu, la fouille de textes ainsi que la simplification lexicale de textes. La désambiguïsation sémantique permet de choisir le sens des unités lexicales polysémiques dans un texte. Elle s'effectue en tenant compte des contextes où un sens peut apparaître (Ide et Véronis, 1998). L'approche la plus directe consiste à estimer la proximité sémantique entre chaque sens candidat et chaque sens de chaque mot appartenant au contexte du mot à désambiguïser. Une application de cette méthode est proposée dans (Pedersen *et al.*, 2005). Le principal problème est la rapide explosion combinatoire qu'elle engendre (complexité exponentielle). En d'autres termes, si le sens sélectionné d'un mot $w$ dans une combinaison $C$ est $S\_w$ et $T$ une liste de mots appartenant au contexte du mot polysémique à désambiguïser alors le score de combinaison est $Score\ (c) = \sum_{wi \in T} \sum_{wi \in T} sim\ (S\_wi,\ S\_wj)$ et il y a en tout $\prod_{w \in T} N_w$ combinaisons à évaluer, avec $N_w$ le nombre de sens du mot *w*. Par exemple, pour une phrase de 10 mots avec 10 sens en moyenne, il y aurait $10^{10}$ combinaisons possibles. Considérons la phrase suivante tirée du corpus d'évaluation que nous décrivons par la suite dans l'article, [*Place flat palms on either side of the head a_few inches away from the ears, fingers pointing toward the shoulders.*], « place_Verbe » a 16 sens selon le réseau sémantique BabelNet (Navigli et Ponzetto, 2012), « flat_Adj » 15, « palm_Nom » 4, « side_Nom » 15, « head_Nom » 40, « a few_Adj » 1, « inch_Nom » 2, « away_Adj » 3, « ear_Nom » 6, « finger_Nom » 4, « point_Verbe » 14 et « shoulder_Nom » 5, il y a alors 5 806 080 000 combinaisons de sens possibles à analyser. Ce calcul exhaustif est donc très compliqué à réaliser dans des conditions réelles et rend impossible l'utilisation d'un contexte de taille importante.

Dans cet article, nous utilisons deux approches totalement différentes à base d'analyse distributionnelle permettant à la fois de réduire le nombre de combinaisons à évaluer et de ne pas perdre de la cohérence au niveau de la désambiguïsation, voire même de l'améliorer. Baroni et Lenci (2010) ont proposé un travail de synthèse sur les procédures relatives au calcul distributionnel. La clé de notre méthode de désambiguïsation est la sélection des voisins distributionnels les plus proches pour chaque mot polysémique dans le texte. Un travail proche du nôtre est proposé dans (McCarthy *et al.*, 2004). La première approche consiste à réaliser une analyse syntaxique en dépendances permettant d'extraire un ensemble de traits syntaxiques pour chaque mot analysé suivant la méthode de Lin (1998). Cette méthode vise à déterminer la similarité distributionnelle entre un mot polysémique et l'un de ses voisins, en se référant aux traits syntaxiques partagés. La deuxième approche consiste à utiliser un modèle de représentation vectorielle continue des mots (*Word2vec*) dans un espace à *n* dimensions. Nous utilisons le modèle proposé par (Mikolov *et al.*, 2013). La similarité consiste ici à comparer le vecteur du mot polysémique et le vecteur de chacun de ses voisins.

Cet article est organisé comme suit. La section 2 présente un état de l'art des différentes méthodes et travaux de désambiguïsation sémantique. L'approche de désambiguïsation sémantique fondée sur des méthodes d'analyse distributionnelle ainsi que les données de travail sont présentées dans la section 3. La section 4 présente les expériences ainsi que les résultats obtenus avant de conclure dans la section 5.

## 2. Travaux antérieurs

Il existe plusieurs méthodes de désambiguïsation sémantique, deux catégories majoritaires peuvent être distinguées. La première rassemble des systèmes supervisés et repose sur l'utilisation d'un corpus d'apprentissage réunissant des exemples d'instances désambiguïsées de mots (Bakx, 2006 ; Navigli, 2009). La deuxième rassemble des systèmes non supervisés et utilise des connaissances provenant de réseaux sémantiques (Tchechmedjiev, 2012, Lafourcade, 2011, Navigli, 2009). Il existe une autre catégorie de systèmes non supervisés permettant l'exploitation des résultats de méthodes d'acquisition automatique de sens. Dans cet article, nous nous intéressons uniquement aux méthodes reposant sur l'utilisation d'un système de désambiguïsation à base de connaissances.

Plusieurs campagnes d'évaluation ont été organisées pour évaluer la performance des algorithmes de désambiguïsation : Senseval-1 (Kilgarriff et Rosenzweig, 2000), Senseval-2 (Edmonds, 2002), Senseval-3 (Mihalcea et Edmonds, 2004) pour l'anglais et RomansEval, désambiguïsation sémantique des sens pour des langues romanes telles que le français et l'italien (Segond, 2000; Calzolari et Corazzari, 2000). La suite des travaux de désambiguïsation a été explorée dans des campagnes successives qui ont eu lieu tous les trois ans entre 1998 et 2010 et annuellement depuis 2012. Par exemple, SemEval-2007 (Navigli et *al.*, 2007), SemEval-2013 (Navigli *et al.*, 2013) et SemEval-2015 (Moro et Navigli, 2015).

L'un des obstacles majeurs d'une désambiguïsation sémantique pour atteindre de bons résultats est la granularité fine des inventaires de sens. Dans Senseval-3, les systèmes ayant participé à la tâche *English All Words* (EAW) ont atteint une performance autour de 65% (Snyder et Palmer, 2004) avec WordNet (Fellbaum, 1998). Ce dernier a été adopté comme inventaire de sens. Une performance de 72,9% a été obtenue sur la tâche *English Lexical Sample* (ELS). Malheureusement, WordNet est une ressource possédant une granularité fine dont la distinction des sens est difficile à reconnaître par les annotateurs humains (Edmonds et Kilgarriff, 2002). Une désambiguïsation avec un inventaire de sens à granularité forte a alors été proposée dans SemEval-2007 sur les mêmes tâches de Senseval-3 (EAW et ELS). Les résultats ont été meilleurs : 82-83% pour EAW et 88,7% pour ELS. Cela montre que la représentation des sens des unités lexicales a un impact décisif lorsqu'on souhaite atteindre des performances dans les 80-90%. La granularité de l'inventaire de sens est également décisive.

## 3. Méthodologie

Nos méthodes de désambiguïsation sémantique prennent en considération des critères distributionnels. Les expériences que nous avons menées ont été réalisées sur un corpus en anglais. Nous avons déjà mené une première expérience pour le français en utilisant seulement l'approche distributionnelle à base de traits syntaxiques (Billami, 2015), le corpus d'évaluation était de petite taille (6 235 occurrences de mots) et il était difficile de tirer des conclusions sur les résultats obtenus. Les expériences que nous présentons dans cet article sont fondées non seulement sur une seule approche distributionnelle mais aussi sur un corpus d'évaluation de taille importante nous permettant ainsi de valider notre approche.

### *3.1. Données de travail*

### *3.1.1. Corpus de travail*

Nous utilisons le corpus Europarl[1], *European Parliament Proceedings Parallel Corpus* (Koehn, 2005). Dans le but de valider nos résultats sur le français, nous avons choisi ce corpus car il s'agit d'un corpus parallèle. Pour l'anglais, Europarl contient plus de 2 millions de phrases (2 218 201) et près de 54 millions de mots (53 974 751). Nous utilisons Mate-Tools[2] (Bohnet et Nivre, 2012) pour la lemmatisation du corpus, l'annotation en parties du discours ainsi que pour l'extraction des dépendances syntaxiques. Le système utilisé permet de coupler l'étiquetage morphosyntaxique et l'analyse de dépendances avec des arbres non projectifs. Les modèles que nous utilisons sont entraînés sur les données de *CoNLL shared task* 2009[3] (Hajič *et al.*, 2009).

---

[1] http://www.statmt.org/europarl, nous utilisons la version 7 du corpus.

[2] https://code.google.com/archive/p/mate-tools

[3] http://ufal.mff.cuni.cz/conll2009-st

*3.1.2. Corpus d'évaluation*

Le test et l'évaluation de notre méthode portent sur le corpus SemCor (Miller *et al.*, 1993). Il s'agit du corpus le plus connu, le plus grand et le plus largement utilisé en termes de sens étiquetés manuellement avec WordNet. Il contient près de 250 000 occurrences de mots contenues dans 352 textes, dont 80% proviennent du Corpus Brown et 20% proviennent d'un roman « *The Red Badge of Courage* » pour lesquels les mots pleins[4] ont été étiquetés avec WordNet. Au total, SemCor contient 11 860 paragraphes et 37 176 phrases. Tous les mots sont lemmatisés et annotés en parties du discours dans le corpus (le traitement avec Mate-Tools se révèle, de ce fait, innécessaire).

*3.1.3. Ressource lexicale BabelNet*

BabelNet[5] (Navigli et Ponzetto, 2012) est un réseau sémantique multilingue permettant de fournir des sens de mots. Nous avons choisi d'utiliser BabelNet comme base de connaissances au lieu de WordNet parce qu'il propose plus d'informations sur les sens provenant de différentes ressources (Wikipédia, Wiktionnaire, Wikidata, Omega wiki, Open Multilingual WordNet) y compris WordNet.

Comme BabelNet permet d'avoir les correspondances de ses sens avec ceux de WordNet et comme SemCor utilise WordNet comme inventaire de sens, nous avons choisi d'adopter BabelNet pour marquer les étiquettes des sens de mots dans SemCor. Un avantage qu'offre BabelNet est qu'il permet de différencier un concept d'une entité nommée. Dans ce travail, nous considérons les sens comme étant des concepts et nous ne tenons pas compte de la présence des entités nommées. BabelNet propose un *mapping* avec les sens de la version 3.0 de WordNet, nous avons donc utilisé la version 3.0 de SemCor. Malheureusement, certains mots sont annotés avec des sens provenant de la version 1.6 de WordNet et n'ont pas une correspondance avec les sens de BabelNet. Il s'agit de 1 728 sens uniques provenant de WordNet 1.6 correspondant à 8 721 occurrences de mots étiquetées. En termes d'annotation sémantique avec BabelNet, nous avons à disposition de 25 881 sens uniques correspondant à 225 415 occurrences de mots annotés dont seulement 224 370 sont annotés comme étant des concepts. Parmi ces 224 370 occurrences, 699 occurrences sont annotées avec plus d'un sens. Nous avons une proportion de 96,28% de mots annotés manuellement dans SemCor, couverts par BabelNet, et de 95,83% de cas que nous traitons sur l'ensemble des mots annotés.

*3.2. Méthodes d'analyse distributionnelle*

Ces méthodes permettent de mesurer la similarité distributionnelle indiquant le degré de cooccurrence entre un mot cible et son voisin apparaissant dans des contextes similaires. Plus la similarité distributionnelle entre un mot à désambiguïser et ses voisins est forte plus la probabilité d'avoir le sens le plus probable est grande.

*3.2.1. Analyse syntaxique en dépendances*

Nous utilisons la méthode proposée par Lin (1998) à partir des dépendances syntaxiques extraites automatiquement depuis notre corpus de travail. Ces dépendances sont stockées et indexées. Nous avons à disposition un ensemble de relations grammaticales de dépendances syntaxiques. Cet ensemble nous permet de mesurer le degré de cooccurrence entre deux mots. La seule condition pour que la méthode fonctionne est que les deux mots doivent partager la même partie du discours. Prenons l'exemple suivant : « *we 've moved on* ». Les triplets[6] de dépendances syntaxiques retournés par Mate-Tools sont : *(move, sbj, we), (move, aux, have), (move, advmod, on) et (move, punct, .)*. Nous pouvons voir les triplets comme des traits syntaxiques : pour le triplet *(move, advmod, on)*, *on* a pour trait syntaxique *advmod (move)*. La similarité distributionnelle entre deux mots est définie par la fonction suivante :

$$sim(w_1, w_2) = \frac{2x\ I(F(w_1) \cap F(w_2))}{I(F(w_1)) + I(F(w_2))}$$

---

[4] Mots pleins : noms, verbes, adjectifs et adverbes.

[5] Nous utilisons la version 2.5.1 de cette ressource lexicale. http://babelnet.org

*F ($w_1$) et F ($w_2$)* représentent l'ensemble des traits syntaxiques possédés respectivement par $w_1$ et $w_2$. $F(w_1) \cap F(w_2)$ représente l'ensemble des traits syntaxiques communs entre $w_1$ et $w_2$. Si *I(S)* est la quantité d'information contenue dans l'ensemble des traits de S alors $I(S) = -\sum_{f \in S} \log P(f)$ où *P(f)* est la probabilité d'avoir le trait syntaxique *f*. Cette similarité prend une valeur entre 0 et 1. Elle retourne 1 si $w_1$ et $w_2$ partagent les mêmes traits syntaxiques et retourne 0 si aucun trait syntaxique de $w_1$ n'est partagé avec les traits syntaxiques de $w_2$. La probabilité *P(f)* est estimée par le pourcentage des mots possédant le trait syntaxique *f* parmi l'ensemble des mots ayant la même partie du discours du mot analysé.

*3.2.2. Word2vec*

Les représentations vectorielles continues des unités lexicales sont en plein essor et ont déjà été appliquées avec succès à de nombreuses tâches en traitement automatique de la langue (Sahlgren, 2008 ; Baroni et Lenci, 2010 ; Mikolov *et al.*, 2013). Il s'agit de projeter les mots selon un modèle de langage dans un espace dans lequel les relations sémantiques entre ces mots peuvent être observées ou mesurées. La technique des Word2vec (Mikolov *et al.*, 2013) construit un réseau de neurones dont le but est de projeter les mots d'une langue (contenus dans une fenêtre sémantique définie) dans un espace de représentation vectorielle. Chaque mot est représenté par un vecteur plein, de taille modérée, qui correspond à une projection du mot dans un espace où les distances modélisent les relations inter-mots. Cette projection permet de tirer profit des mots selon leurs sens dans une région de l'espace sémantique proche. Par exemple, « *Paris* » et « *London* » peuvent partager l'idée de « *capital city* ».

Il y a deux types de modèles Word2vec : le premier repose sur une architecture fondée sur les sacs-de-mots continus (*continuous bag of words* ou CBOW), le deuxième repose sur une architecture fondé sur les Skip-Grams. Ces architectures sont manipulées par un réseau de neurones. Le modèle CBOW cherche à prédire un mot selon son contexte alors que le modèle Skip-Grams cherche à prédire un contexte sachant un mot. Dans notre travail, nous utilisons le modèle Skip-Grams parce que nous nous intéressons à la sélection d'un contexte réduit en terme de taille et permettant de retourner un certain nombre *k* des mots les plus pertinents par rapport au contexte d'origine. Nous utilisons une alternative du projet Word2vec[7] fait en java par l'équipe Medallia[8] pour l'intégrer dans notre programme principal, fait en java, de désambiguïsation sémantique. L'entraînement du réseau de neurones est réalisé sur les mêmes données du corpus de travail avec un prétraitement. Tous les mots sont lemmatisés et annotés en parties du discours avec la chaîne de traitement Mate-Tools. Par exemple, la phrase « *In short, the issue is an important one.* » est remplacé par « *in_IN short_Adj ,_, the_DT issue_N be_V an_DT important_Adj one_N ._.* ». Pour le paramétrage du réseau de neurones, nous avons choisi une fenêtre d'une taille de 20 mots (leur fréquence d'apparition est d'un minimum égal à 5, la dimension des vecteurs est de 300, le nombre d'exemples négatifs est de 7 avec une utilisation de l'alternative « *softmax* hiérarchique »). Pour mesurer la similarité entre mots, nous utilisons la mesure *cosinus*. Cette méthode possède l'avantage de ne pas être dépendante de la partie du discours des mots à comparer.

---

[6] Un triplet de dépendance syntaxique se compose d'un nom de la relation, d'un gouverneur et d'un dépendant.

[7] https://github.com/medallia/Word2VecJava

[8] http://engineering.medallia.com

### 3.3. Similarités sémantiques

Nous utilisons l'algorithme de Lesk (1986) et ses variantes pour mesurer la similarité sémantique. Ces algorithmes nécessitent un dictionnaire (BabelNet dans notre cas) et aucun apprentissage. L'algorithme de base de Lesk est très simple : il considère la similarité entre deux sens comme le nombre de mots pleins en commun dans leurs définitions que nous appelons par la suite *« traits sémantiques »*, sans tenir compte de l'ordre des mots. Nous utilisons Mate-Tools pour obtenir ces traits sémantiques. La fonction utilisée pour mesurer la similarité sémantique se présente par :

$$Sim_{Lesk}(S_1, S_2) = |D(S_1) \cap D(S_2)|$$

Dans le cas où aucune définition n'est proposée pour un sens, nous tenons compte des synonymes. D'un autre côté, nous utilisons une variante de l'algorithme de Lesk (Navigli, 2009) consistant à comparer chaque sens candidat avec le contexte du mot *w* à désambiguïser. Comme contexte, nous tenons compte de la phrase dont laquelle le mot polysémique apparaît. La fonction utilisée se présente par :

$$Lesk_{Variante} = |contexte(w) \cap D(S_i(w))|$$

$S_i$ est l'IIIème sens du mot *w*. L'algorithme de Lesk est très sensible à la présence des mots. Une absence des mots représentant fortement les sens dans les définitions retourne des résultats qui ne sont pas de bonne qualité. Nous utilisons l'algorithme de Lesk étendu (Banerjee et Pedersen, 2002) pour faire face à cette limite. Nous avons préféré d'utiliser une version simplifiée[9] de l'algorithme au lieu de la version originale pour des raisons calculatoires. Cette version simplifiée consiste à comparer les traits sémantiques dans les définitions des sens des mots du contexte ainsi que dans les définitions des sens provenant de différentes relations telles que *l'hyperonymie, l'hyponymie, la méronymie* ou *l'holonymie*.

### 3.4. Désambiguïsation

Nous pouvons voir notre méthode comme un processus à deux niveaux : le premier sélectionne les voisins les plus proches au moyen d'une similarité distributionnelle, le deuxième permet de lever les ambiguïtés au moyen d'une similarité sémantique. La similarité distributionnelle entre le mot à désambiguïser et chacun des voisins sélectionnés est plus forte que celle du mot à désambiguïser et chacun des autres mots du contexte. La similarité sémantique utilisée tient compte des traits sémantiques provenant des définitions des sens. Le sens candidat choisi pour un mot polysémique est censé être celui qui partage le plus de traits sémantiques avec les sens des voisins sélectionnés. A titre d'exemple, le sens *lawyer* (*homme de loi*) du mot *lawyer* partage plus de traits sémantiques avec le sens *law* (*loi*) du mot *law* que le sens *lawyer* (*poisson*) peut partager. Nous adaptons une méthode structurelle fondée sur une distance sémantique entre sens selon la formule proposée par (Navigli, 2009) :

$$S^* = argmax_{S \in \text{Sens}(w)} \sum_{N_i \in N_w : N_i \neq w} max \quad \text{Score}(S, S')$$

---

[9] Nous faisons une comparaison entre des unités lexicales (mots) et non pas sur des séquences de mots.

$S' \in \text{Sens}(N_i)$ $avec\ i = 1 \ldots k$ et $N_w = \{N_1, N_2, \ldots, N_k\}$ est l'ensemble ordonné des $k$ voisins les plus proches du mot cible $w$. $\text{Sens}(N_i)$ est l'ensemble des sens du voisin $N_i$ et $\text{Sens}(w)$ est l'ensemble des sens du mot cible $w$. $\text{Score}(S, S')$ est la fonction utilisée pour mesurer la similarité entre deux sens $S$ et $S'$. Nous utilisons la phrase, dans laquelle le mot à désambiguïser apparaît, comme contexte. Dans le cas d'une égalité de *score* entre plusieurs sens candidats, nous utilisons une heuristique qui tient compte du sens possédant le plus grand nombre de connexions sémantiques dans le réseau BabelNet.

## 4. Expériences

Dans Senseval-1 et Senseval-2, des variantes de l'algorithme de Lesk ont été considérées soit comme des approches de base soit comme des systèmes complets. Dans Senseval-1, la plupart des systèmes de désambiguïsation ayant participé à la tâche All-Words (AW) ont été surclassés par une variante de Lesk (Kilgarriff, Rosenzweig, 2000). D'un autre côté, durant Senseval-2, les algorithmes à base de Lesk ont été surclassés par la plupart des systèmes ayant participé à la tâche Lexical-Sample.

Notre système retourne le même nombre de réponses que les données sur lesquelles nous prenons une référence (224 370 occurrences de mots dont 191 146 occurrences sont pour des mots polysémiques). Afin de mesurer la performance de notre système de désambiguïsation, nous ne tenons pas compte des mots où BabelNet renvoie un seul sens candidat. Nous utilisons le taux d'exactitude pour mesurer cette performance.

### 4.1. Protocole expérimental

Pour comparer les performances de notre approche, nous avons choisi d'une part de faire des expériences sur l'ensemble des mots polysémiques du corpus d'évaluation et, d'autre part, de faire des expériences sur un échantillon de mots polysémiques. Notre évaluation se porte sur la sélection des voisins distributionnels ainsi que sur l'algorithme de désambiguïsation. Nous faisons une comparaison avec les voisins les plus proches linéairement en les sélectionnant de la droite vers la gauche. Nous utilisons un paramètre $k$ pour choisir le nombre de voisins à sélectionner. Le choix se porte sur une valeur entre 2 et 7. Nous avons présenté dans la sous-section 3.2 deux mesures distributionnelles, la première repose sur la méthode de Lin et la deuxième sur la méthode Word2vec (W2V). Dans nos expériences, nous utilisons une autre mesure distributionnelle (ALL) représentant une combinaison[10] des deux mesures.

### 4.2. Jeu de test

Les mots du jeu de test sont choisis selon leur niveau d'ambiguïté (peu ambigu, ambigu ou très ambigu). Nous avons fait une sélection de quatre mots pour les catégories grammaticales noms, verbes et adjectifs : noms= {*argument, disc, paper, plan*}, verbes= {*operate, note, add, begin*}, adjectifs= {*black, valid, wet, narrow*}. Nous avons à disposition un ensemble de 12 mots représentant 1 022 occurrences dans SemCor. Nous comparons nos résultats avec Babelfy (Moro *et al.*, 2014), un système de désambiguïsation utilisant BabelNet comme base de connaissances.

---

[10] Nous utilisons une moyenne entre les deux mesures et prenons en compte des voisins partageant la même partie du discours avec le mot à désambiguïser.

### *4.3. Résultats*

Le tableau 1 présente les résultats obtenus sur l'ensemble des mots polysémiques traités dans le corpus SemCor (191 146 occurrences de mots). Ce tableau tient compte de toutes les méthodes d'analyse distributionnelle et tous les algorithmes utilisés de désambiguïsation en sélectionnant quatre voisins. Sur ce même tableau, nous présentons une comparaison avec la sélection des voisins les plus proches linéairement. L'algorithme LeskVar fait référence à la variante de Lesk, LeskB-PPVL présente l'application de l'algorithme de base de Lesk sur les voisins les plus proches linéairement, LeskES-PPVL fait référence à l'application de l'algorithme de Lesk étendu simplifié sur ces mêmes voisins, LeskB-PPVD pour l'algorithme de base de Lesk en utilisant les voisins distributionnels et LeskES-PPVD pour l'application de l'algorithme de Lesk étendu simplifié sur ces voisins distributionnels.

| Algorithmes/ POS | LeskVar | LeskB-PPVL | LeskES-PPVL | LeskB-PPVD | | | LeskES-PPVD | | |
|---|---|---|---|---|---|---|---|---|---|
| | | | | Lin | W2V | ALL | Lin | W2V | ALL |
| **Noms** | 40,7% | 40,6% | 45,3% | 41,4% | **40,0%** | 41,3% | **47,3%** | 44,7% | 47,2% |
| **Verbes** | **34,4%** | **25,2%** | 30,0% | 29,9% | 28,0% | 30,0% | 30,2% | 26,1% | 30,2% |
| **Adjectifs** | 48,3% | 45,3% | 44,8% | 48,0% | 43,3% | 47,9% | **49,4%** | **40,4%** | **49,4%** |
| **Adverbes** | 45,9% | **47,6%** | 42,0% | 45,8% | 46,4% | 45,7% | 42,3% | **41,7%** | 42,3% |

*Tableau 1 : Taux d'exactitude obtenus selon différents algorithmes de désambiguïsation (k=4)*

En termes de comparaison des méthodes d'analyse distributionnelle, il s'avère que l'utilisation de la méthode de Lin est plus rentable et meilleure, en qualité, que l'utilisation de la méthode W2V. Nous remarquons aussi que la combinaison des deux méthodes ne retourne pas des résultats aussi meilleurs que la simple utilisation de la méthode de Lin. L'utilisation d'une analyse syntaxique en dépendances pour la tâche de désambiguïsation sémantique reste meilleure. L'algorithme de Lesk étendu simplifié retourne les meilleurs résultats pour les noms et les adjectifs. Pour le cas des verbes, nous pouvons avoir les meilleurs résultats sans aller chercher l'information sémantique dans les relations sémantiques, il suffit de tenir compte des mots du contexte (variante de Lesk). Pour le type des voisins à utiliser (*distributionnel vs linéaire*), le voisin distributionnel rend la désambiguïsation sémantique encore meilleure que le voisin le plus proche linéairement et cela pour les noms, verbes et adjectifs. Une exception est à noter pour les adverbes où les résultats sont plus ou moins proches selon l'algorithme utilisé.

La figure 1 présente l'application de l'algorithme de Lesk étendu simplifié dont le nombre de voisins est variable. Nous remarquons que l'utilisation d'un petit nombre de voisins distributionnels peut nous mener à atteindre les meilleurs résultats. La figure 2 présente les résultats seulement sur les occurrences de mots annotées manuellement avec le sens le plus fort dans BabelNet. Nous aurions pu imaginer avec l'heuristique utilisée qu'on peut atteindre des résultats dans les 80-90% mais ce n'est pas le cas. Nous atteignons seulement 75% pour les noms et 67% sur l'ensemble des mots avec une utilisation de la méthode de Lin (*cf.* figure 2). BabelNet comme WordNet et comme toute autre ressource lexicale contient des erreurs et des incohérences et celles-ci se traduisent souvent par des anomalies dans SemCor.

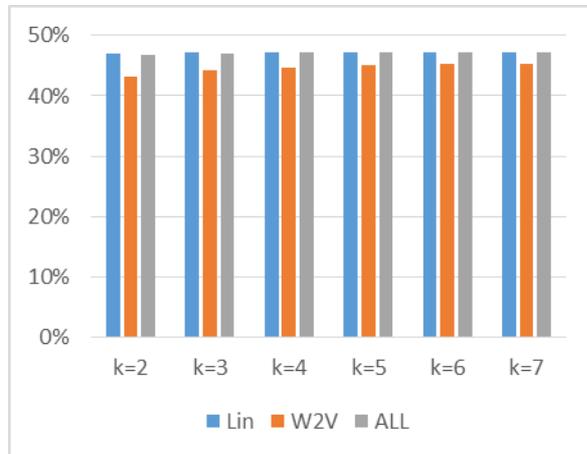

*Figure 1 : Résultats obtenus par l'application de l'algorithme de Lesk étendu simplifié sur l'ensemble des mots polysémiques du corpus SemCor*

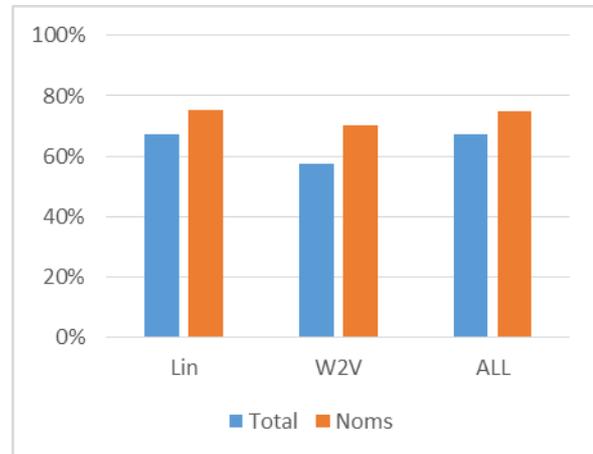

*Figure 2 : Résultats obtenus seulement sur les mots ayant été annotés manuellement par le sens le plus fort selon BabelNet, application de l'algorithme de Lesk étendu simplifié avec k=4*

La figure 3 présente les résultats sur les mots du jeu de test par application de la méthode de Lin. La figure 4 présente les résultats sur ces mêmes mots par application des différentes méthodes d'analyse distributionnelle et selon l'algorithme LeskES-PPVD. A ce stade, le meilleur taux d'exactitude est de 58% contre 61% seulement pour Babelfy.

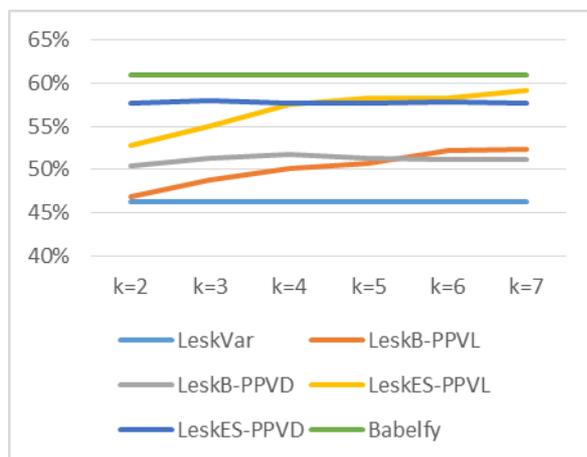

*Figure 3 : Résultats obtenus suivant une analyse syntaxique en dépendances et une comparaison avec Babelfy sur l'ensemble des mots du jeu de test*

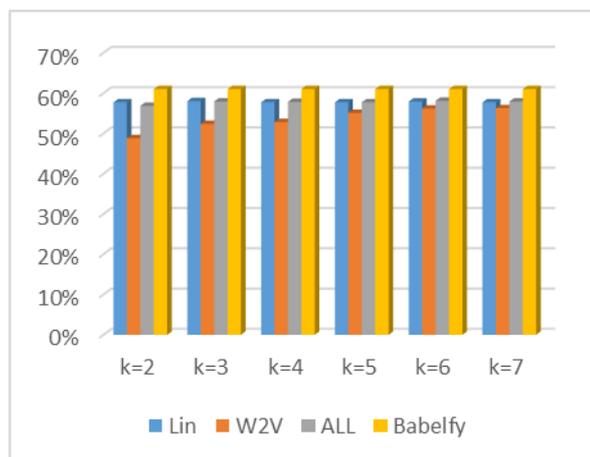

*Figure 4 : Résultats obtenues selon une utilisation de différentes approches à base d'analyse distributionnelle sur l'ensemble des mots du jeu de test et selon une application de l'algorithme de Lesk étendu simplifié, de plus une comparaison avec Babelfy*

## 5. Conclusion

Cet article propose une méthode permettant de réduire la complexité exponentielle qu'engendre l'algorithme le plus direct de désambiguïsation sémantique. Nous avons remarqué que l'utilisation des voisins distributionnels permet non seulement de réduire cette complexité mais aussi de garder une cohérence au niveau de la désambiguïsation. Pour lever l'ambiguïté des noms et des adjectifs, l'utilisation des voisins distributionnels reste meilleure que la simple utilisation des voisins les plus proches linéairement. Nous prévoyons, en perspectives, d'étendre cette étude en travaillant sur d'autres similarités sémantiques que la simple comparaison par égalité des traits sémantiques entre sens afin d'augmenter les performances de notre système. Nous pouvons utiliser les mesures de similarité distributionnelles proposées pour calculer une proximité sémantique entre sens et cela en tenant compte de chaque paire possible de mots.

## Références

Bakx G. E. (2006). Machine Learning Techniques for Word Sense Disambiguation. *Dept. LSI*. Universitat Politècnica de Catalunya, Barcelona, Catalunya.

Banerjee S. et Pedersen T. (2002). An adapted lesk algorithm for word sense disambiguation using wordnet. *In CICLing*, London, UK, pages 136–145.

Baroni M. et Lenci A. (2010). Distributional memory: A general framework for corpus based semantics. *Computational Linguistics*, **36**(4), 673–721.

Billami M. B. (2015). Désambiguïsation lexicale à base de connaissances par sélection distributionnelle et traits sémantiques. Actes de la 22e conférence en Traitement Automatique des Langues Naturelles, session RECITAL, Caen, pages 13–24.

Bohnet B. et Nivre J. (2012). A Transition-Based System for Joint Part-of-Speech Tagging and Labeled Non-Projective Dependency Parsing. *EMNLP-CoNLL*, pages 1455–1465.

Calzolari N. et Corazzari O. (2000). Senseval/Romanseval: The Framework for Italian, Computers and the Humanities 34: 61–78, Kluwer Academic Publishers, Printed in Netherland.

Edmonds P. (2002). SENSEVAL: The Evaluation of Word Sense Disambiguation Systems, ELRA Newsletter, Vol. 7, No. 3.

Edmonds P. et Kilgarriff A. (2002). Introduction to the special issue on evaluating word sense disambiguation systems. *Journal of Natural Language Engineering*, **8**(4):279–291.

Fellbaum C. (1998). WordNet: an Electronic Lexical Database. MIT Press.

Hajič J., Ciaramita M., Johansson R., Kawahara D., Marti M. A., Màrquez L., Meyers A., Nivre J., Padó S., Štěpánek J., Straňák P., Surdeanu M., Xue N. et Zhang Y. (2009). The conll-2009 shared task: Syntactic and semantic dependencies in multiple languages. *In Proceedings of the 2009 CoNLL Shared Task*, pages 1–18.

Ide N. et Véronis J. (1998). Word sense disambiguation: The state of the art. *Computat. Ling.* **24**(1), 1–41, MIT Press, Cambridge, MA, USA.

Kilgarriff A. et Rosenzweig J. (2000). Framework and Results for English SENSEVAL, *Computers and the Humanities*, 34, pages 15–48.

Koehn P. (2005). Europarl: A Parallel Corpus for Statistical Machine Translation, MT Summit 2005.

Lafourcade M. (2011). Lexique et analyse sémantique de textes – structures, acquisition, calculs et jeux de mots. *Mémoire d'habilitation à diriger les recherches*, Université Montpellier 2, LIRMM. Soutenu le 7 décembre 2011.

Lesk M. (1986). Automatic sense disambiguation using machine readable dictionaries: How to tell a pine cone from an ice cream cone. *In Proceedings of the 5th SIGDOC*, New York, pages 24–26.

Lin D. (1998). An information-theoretic definition of similarity. *In Proceedings of the 15th International Conference on Machine Learning* (ICML, Madison, WI), pages, 296–304.

Mccarthy D., Koeling R., Weeds, J. et Carroll J. (2004). Finding predominant senses in untagged text. *In Proceedings of the 42nd Annual Meeting of the Association for Computational Linguistics*, Barcelona, Spain, pages 280–287.

Mihalcea R., Edmonds P. (2004). Proceedings of the 3rd International Workshop on the Evaluation of Systems for the Semantic Analysis of Text (Senseval-3, Barcelona, Spain).

Mikolov T. et Corrado G., Chen K. et Dean J. (2013). Efficient Estimation of Word Representations in Vector Space. *Proceedings of the International Conference on Learning Representations* (*ICLR* 2013), pages 1–12.

Miller G. A., Leacock C., Tengi R. et Bunker R. T. (1993). A semantic concordance. *In Proceedings of the ARPA Workshop on Human Language Technology*, pages 303–308.

Moro A. et Navigli R. (2015) SemEval-2015 Task 13: Multilingual All-Words Sense Disambiguation and Entity Linking. *Proceedings of the 9th International Workshop on Semantic Evaluation (SemEval), in the 2015 Conference of the North American Chapter of the Association for Computational Linguistics (NAACL 2015)*, Denver, Colorado, June 4-5th, 2015, pages 288–297.

Moro A., Raganato A. et Navigli R. (2014). Entity Linking meets Word Sense Disambiguation: a Unified Approach. *Transactions of the Association for Computational Linguistics (TACL)*, 2, pages 231–244.

Navigli R. (2009). Word Sense Disambiguation: a Survey. *ACM Computing Surveys*, **41**(2), ACM Press, 2009, pages 1–69.

Navigli R., Jurgens D. A. et Vannella D. (2013). SemEval-2013 Task 12: Multilingual Word Sense Disambiguation. *Proceedings of 7th International Workshop on Semantic Evaluation (SemEval), in the Second Joint Conference on Lexical and Computational Semantics*, Atlanta, USA, June 14-15th, 2013, pages 222–231.

Navigli R., Litkowski K. et Hargraves O. (2007) SemEval-2007 Task 07: Coarse-Grained English All-Words Task. *Proceedings of Semeval-2007 Workshop (SEMEVAL), in the 45th Annual Meeting of the Association for Computational Linguistics (ACL 2007)*, Prague, Czech Republic, June 23-24th, 2007, pages 30–35.

Navigli R. et Ponzetto S. (2012). BabelNet: The Automatic Construction, Evaluation and Application of a Wide-Coverage Multilingual Semantic Network. *Artificial Intelligence*, 193, Elsevier, 2012, pages 217–250.

Pedersen T., Banerjee S. et Patwardhan S. (2005). Maximizing Semantic Relatedness to Perform Word Sense Disambiguation. *Research Report UMSI 2005/25*, University of Minnesota Supercomputing Institute.

Sahlgren M. (2008). The distributional hypothesis. *Italian Journal of Linguistics*, **20**(1), 33–54.

Segond F. (2000). Framework and Results for French, Computers and the Humanities 34: 49–60, Kluwer Academic Publishers, Printed in Netherland.

Snyder B. et Palmer. M. (2004). The English All-Words Task. *In Proceedings of SENSEVAL-3*, pages 41–43.

Tchechmedjiev A. (2012). État de l'art : mesures de similarité sémantique locales et algorithmes globaux pour la désambiguïsation lexicale à base de connaissances. *In Proceedings of the Joint Conference JEP-TALN-RECITAL 2012*, volume 3: RECITAL, ATALA/AFCP. June 2012, Grenoble, France, pages 295–308.